\pdfoutput=1
\documentclass[letterpaper]{article}
\usepackage[preprint]{aaai2027}
\usepackage[hyphens]{url}
\usepackage{graphicx}
\usepackage{natbib}
\usepackage{caption}
\usepackage{booktabs}
\usepackage{amsmath}
\usepackage{amssymb}
\usepackage{amsfonts}

\newcommand{\R}{\mathbb{R}}
\newcommand{\eps}{\epsilon}
\newcommand{\ginf}{\gamma_{\infty}}
\newcommand{\uk}{u_{\kappa}}
\newcommand{\one}{\mathbf{1}}

\newcommand{\Rdyn}{R_{\mathrm{dyn}}}
\newcommand{\dist}{\operatorname{dist}}
\newcounter{proposition}
\newcommand{\propheading}[1]{\refstepcounter{proposition}\noindent\textbf{Proposition \theproposition\ (#1).}}
\newcounter{theorem}
\newcommand{\theoremheading}[1]{\refstepcounter{theorem}\noindent\textbf{Theorem \thetheorem\ (#1).}}

\title{Stability Annealing Selects the Implicit Bias of Smoothed Sign Descent: A Rate-Indexed Barrier Path on Separable Data}
\author{Xiangwu Wang\textsuperscript{1}, Chengwei Cao\textsuperscript{2}, Yicheng Song\textsuperscript{3}, Ran Bi\textsuperscript{1}, Peilin Yu\textsuperscript{1}}
\affiliations{\textsuperscript{1}The University of Hong Kong\\
\textsuperscript{2}University of California, San Diego\\
\textsuperscript{3}Beijing University of Aeronautics and Astronautics}

\begin{document}

\maketitle

\begin{abstract}
Adaptive gradient methods can favor max-margin separators that differ from gradient descent, yet a fixed positive numerical stability constant eventually changes the update geometry again. This paper studies the rate-controlled middle case for full-batch linear classification on separable data. For memoryless stability-annealed smoothed-sign descent with weighted exponential loss, we prove that the normalized iterates converge to the minimizer of a convex Burg-type barrier over a margin slice. The proof rewrites the dynamics exactly as entropic mirror ascent on a concave dual objective, controls the dual gap by a KL recursion, and yields an explicit \(S_t^{-1/2}\) normalized-iterate envelope. The static barrier geometry is fully characterized, including KKT conditions and both endpoint limits. Experiments validate the exact dual identities to floating-point error, illustrate the predicted path and rate diagram, and show an empirical fixed-\(\epsilon\) crossover scaling in cumulative time. We further report robustness and boundary diagnostics for logistic tails, fixed-\(\epsilon\) crossover, and adaptive-method variants, delineating the scope of the proved smoothed-sign theory.
\end{abstract}

\section{Introduction}

Implicit bias results for separable classification usually compare endpoints: gradient descent converges in direction to an \(\ell_2\)-margin separator, while sign-like, normalized, and adaptive methods are linked to non-Euclidean or \(\ell_\infty\)-type geometry \citep{soudry2018implicit,nacson2019stochastic,lyu2020gradient,kingma2015adam,wilson2017marginal,zhang2024implicitadam,fan2025spectral}. The numerical stability constant in Adam-like updates complicates that endpoint story. When the stability term is absent or negligible, the update resembles a sign direction. When it is fixed and positive, sufficiently small gradients eventually see a more gradient-descent-like denominator. The motivating contradiction is that both endpoints are plausible, but neither specifies what separator is selected when the stability term is deliberately annealed at a prescribed exponential rate.

\begin{figure}[t]
\centering
\includegraphics[width=\columnwidth]{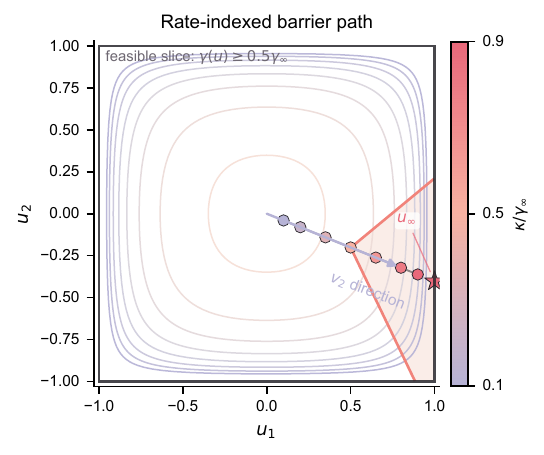}
\caption{Two-dimensional barrier geometry. Contours show the Burg-type barrier, the shaded region is one feasible margin slice, markers are \(\uk\) colored by \(\kappa/\ginf\), the arrow shows the small-\(\kappa\) Euclidean direction, and the star is the \(\ell_\infty\)-margin endpoint. The selected dataset is illustrative only and is excluded from aggregate statistics.}
\label{fig:concept}
\end{figure}

We study this question first for the memoryless smoothed-sign proxy
\begin{equation}
w_{t+1}=w_t+\eta_t
\frac{-\nabla L(w_t)}{|\nabla L(w_t)|+\eps_t},
\qquad
\eps_t=\eps_0\exp(-\kappa S_t),
\label{eq:softsign}
\end{equation}
where \(S_t=\sum_{s=0}^{t-1}\eta_s\), and absolute values and divisions are coordinatewise. Equation~\eqref{eq:softsign} removes Adam's exponential moving averages while retaining the coordinatewise competition between gradient magnitude and stability. The central object is not a generic transition curve. It is the rate-indexed constrained barrier path shown in Figure~\ref{fig:concept}.

This paper makes four contributions. First, it gives an exact rate-indexed implicit-bias theorem for stability-annealed smoothed-sign descent under weighted exponential loss. Second, it identifies the limiting separator as a Burg-barrier minimizer over a margin slice and characterizes the endpoint geometry. Third, it proves convergence through an exact entropic mirror-ascent representation and a KL recursion. Fourth, it reports theorem-compatible numerical diagnostics, including floating-point dual residuals and a long-horizon showcase, together with boundary diagnostics for logistic tails, fixed-\(\epsilon\) crossover, and adaptive-method variants. The scope is precise: the theorem is for exponential-loss smoothed-sign dynamics, logistic behavior is empirical, the critical rate \(\kappa=\ginf\) remains outside Theorem~\ref{thm:dynamic}, and Adam/RMSProp are boundary diagnostics rather than theorem claims.

\section{Related Work and Problem Setup}

For separable linear classification, exponential or logistic tails make unregularized gradient descent diverge in norm while its direction approaches the \(\ell_2\) hard-margin separator \citep{soudry2018implicit,nacson2019convergence}; boosting and margin-path analyses connect such tails, shrinkage, and relaxed constraints to margin maximization \citep{rosset2004boosting,telgarsky2013margins}. Nonseparable and regularization-path results extend this picture for general decreasing losses \citep{ji2019nonseparable,ji2020regularization}. We use this literature for the baseline lesson that exponential tails select margin geometry without explicit regularization.

Which margin geometry appears depends on the update. Steepest- and mirror-descent views tie the separator to the norm or potential defining the algorithm \citep{gunasekar2018geometry,beck2003mirror}, and generalized-margin or mirror-flow analyses make this dependence explicit on separable data \citep{sun2022mirror,pesme2024mirror}. Homogeneous and wide-network analyses likewise show that margin or KKT-type limits can depend on architecture, parameterization, and function-space geometry \citep{nacson2019lexicographic,chizat2020wide,vardi2021margin,lyu2021twolayer}. We use standard convex-duality and Bregman tools \citep{rockafellar1970convex,bregman1967relaxation}; the new object is the rate-indexed selector induced by stability annealing, not a new barrier.

Adaptive and sign-like methods motivate the opposing endpoint. AdaGrad and Adam use coordinatewise scaling \citep{duchi2011adaptive,kingma2015adam}; adaptive methods can generalize differently from SGD, and AMSGrad addresses Adam convergence pathologies \citep{wilson2017marginal,reddi2018convergence}. Sign-based optimization isolates the coordinatewise sign geometry behind our memoryless proxy \citep{bernstein2018signsgd,balles2018dissecting}, while Adam/AdaGrad convergence analyses in smooth bounded-gradient settings address a complementary optimization question \citep{defossez2020simple}. Work on AdaGrad, adaptive methods on homogeneous networks, finite-step Adam/RMSProp, and Adam on separable data shows optimizer-dependent implicit bias, including \(\ell_\infty\)-margin behavior \citep{qian2019adagrad,wang2021adaptive,cattaneo2024adambias,zhang2024implicitadam}. We cite \citet{fan2025spectral} only as normalized/spectral/Muon context, not as an Adam theorem.

Most closely, \citet{wang2025smoothedsign} study smoothed sign descent with a fixed stability constant through a mirror-descent lens, giving dual dynamics and a Bregman/KKT interpretation for fixed-\(\varepsilon\) regression. Our theorem instead studies separable classification with \(\epsilon_t=\epsilon_0e^{-\kappa S_t}\), where \(\kappa\) acts as a margin constraint. The transformed dynamics is exactly entropic mirror ascent on a concave dual objective, and the primal limit is the Burg-barrier minimizer over \(Zu\ge\kappa\mathbf{1}\). The contribution is therefore a rate-indexed implicit-bias theorem, not another fixed-\(\varepsilon\) smoothed-sign interpretation.

Let \(z_i=y_i x_i\in\R^d\) be signed examples. We write
\begin{equation}
\gamma(u)=\min_{i\in[n]}z_i^\top u,
\qquad
\ginf=\max_{\|u\|_\infty\le 1}\gamma(u).
\label{eq:margins}
\end{equation}
Here \(\gamma(w_t)\) is the raw margin of the iterate, while \(\gamma(w_t/S_t)\) and \(\gamma(w_t/\|w_t\|_\infty)\) are normalized margins on different scales. Throughout, \(t\)-time denotes iteration count and \(S_t\)-time denotes cumulative learning-rate time. The experiments compare trajectories primarily in \(S_t\)-time because the stability schedule is defined through \(S_t\).

\noindent\textbf{Static assumptions.}
The barrier program uses only the signed examples \(z_i=y_i x_i\), separability in the \(\ell_\infty\) sense \(\ginf>0\), and an interior rate \(0<\kappa<\ginf\). The margin function and max-margin value are those in \eqref{eq:margins}. The cumulative time and annealed stability are
\begin{equation}
S_t=\sum_{s=0}^{t-1}\eta_s,
\qquad
\eps_t=\eps_0\exp(-\kappa S_t).
\label{eq:time-schedule}
\end{equation}
These assumptions are sufficient for the static convex program and endpoint statements below. They do not assert dynamic convergence of the iterates.

\noindent\textbf{Dynamic theorem assumptions.}
The exact dynamic theorem is stated for weighted exponential loss
\begin{equation}
L(w)=\sum_{i=1}^n a_i e^{-z_i^\top w},
\qquad a_i>0 .
\label{eq:weighted-exp-loss}
\end{equation}
It assumes \(\eta_t>0\), \(S_t\to\infty\), \(\sum_t\eta_t^2<\infty\), and
\begin{equation}
\eta_tG\le1,\qquad
G=\max_i(\kappa+\|z_i\|_1).
\label{eq:eta-g-condition}
\end{equation}
The initialization \(w_0\) is arbitrary but finite. No active-set regularity, residual-prefactor convergence, coordinate noncancellation, bounded normalized iterate assumption, dual multiplier uniqueness, strict complementarity, or generic-position condition is imposed. Logistic loss is evaluated only as an empirical robustness extension unless a separate perturbation theorem is proved.

\noindent\textbf{Procedure.}
The memoryless stability-annealed update used throughout the main experiments is:
\begin{enumerate}
\item choose \(w_0\), stepsizes \(\eta_t>0\), \(\eps_0>0\), and \(0<\kappa<\ginf\);
\item compute \(S_t\) and \(\eps_t\) from \eqref{eq:time-schedule};
\item evaluate the full-batch gradient \(\nabla L(w_t)\);
\item update \(w_{t+1}=w_t+\eta_t(-\nabla L(w_t))/(|\nabla L(w_t)|+\eps_t)\) coordinatewise.
\end{enumerate}

For completeness we also record the full-batch Adam convention tested in the transfer experiment:
\begin{equation}
m_t=\beta_1m_{t-1}+(1-\beta_1)\nabla L(w_t),
\label{eq:adam-moments}
\end{equation}
with second-moment accumulator
\begin{equation}
v_t=\beta_2v_{t-1}+(1-\beta_2)\nabla L(w_t)^{\odot 2},
\label{eq:adam-second}
\end{equation}
with update
\begin{equation}
w_{t+1}=w_t-\eta_t\frac{m_t}{\sqrt{v_t}+\eps_t}.
\label{eq:adam-update}
\end{equation}
Bias-corrected and uncorrected variants are both evaluated. Equations~\eqref{eq:adam-moments}, \eqref{eq:adam-second}, and \eqref{eq:adam-update} are used only for empirical transfer diagnostics in this paper.

The smoothed-sign proxy can also be read as an interpolating coordinate map. If \(|\nabla_j L(w_t)|\gg \eps_t\), the \(j\)th update is close to a sign step. If \(|\nabla_j L(w_t)|\ll \eps_t\), it is close to a scaled gradient step. The annealing rate therefore asks which coordinates cross this comparison at the same exponential scale as the margin tail. The barrier program below is the static object obtained when this coordinatewise comparison is compatible across all active support vectors.

\section{Rate-Indexed Barrier Geometry}

\subsection{Barrier Properties}

For \(u\in(-1,1)^d\), define
\begin{equation}
B(u)=\sum_{j=1}^d\left[-|u_j|-\log(1-|u_j|)\right].
\label{eq:barrier}
\end{equation}
Let \(b(r)=-r-\log(1-r)\) for \(r\in[0,1)\). Then
\begin{equation}
b'(r)=\frac{r}{1-r},
\qquad
b''(r)=\frac{1}{(1-r)^2}.
\label{eq:scalar-barrier-calculus}
\end{equation}
Thus \(b(0)=0\), \(b\) is nonnegative, and \(b\) is increasing and strictly convex away from the origin. The vector barrier is differentiable at zero because the one-sided coordinate derivatives agree there. For every coordinate,
\begin{equation}
[\nabla B(u)]_j=\frac{u_j}{1-|u_j|}.
\label{eq:barrier-gradient}
\end{equation}
Moreover, \(B\) is convex and \(1\)-strongly convex on the open cube in the Euclidean sense. The coordinate second derivative away from zero is \(1/(1-|u_j|)^2\), and the lower bound by \(1\) extends through the origin by convexity.

\subsection{Well-posedness and KKT}

Let \(Z\) be the matrix with rows \(z_i^\top\). The static candidate at rate \(0<\kappa<\ginf\) is
\begin{equation}
\uk=\arg\min_{u\in(-1,1)^d}B(u)
\quad\mathrm{s.t.}\quad
z_i^\top u\ge \kappa\ \ \forall i.
\label{eq:barrier-program}
\end{equation}

\propheading{Well-posedness and KKT}\label{prop:static-kkt}
Under separability with \(\ginf>0\) and \(0<\kappa<\ginf\), program~\eqref{eq:barrier-program} attains a unique minimizer \(\uk\). It satisfies \(\gamma(\uk)=\kappa\), so at least one margin constraint is active. There are multipliers \(\lambda_i\ge0\) such that
\begin{align}
Z\uk&\ge\kappa\one,\qquad \lambda\ge0,\notag\\
\lambda_i(z_i^\top\uk-\kappa)&=0,\qquad
\nabla B(\uk)=Z^\top\lambda .
\label{eq:kkt}
\end{align}
Conversely, any feasible point satisfying \eqref{eq:kkt} is the unique minimizer.

\noindent\textit{Proof idea.}
Let \(u_\infty\) maximize \(\gamma\) over \(\|u\|_\infty\le1\). Since \(0<\kappa<\ginf\), the inward scaling \(\tilde u=\frac12(1+\kappa/\ginf)u_\infty\) is strictly feasible and lies in the open cube. Minimize the extended-value version of \(B\), equal to \(+\infty\) on the boundary, over the compact closed feasible set \(\{u:\|u\|_\infty\le1, Zu\ge\kappa\one\}\). The strict feasible point makes the infimum finite, and divergence at \(|u_j|=1\) forces any optimizer into the open cube. Strict convexity gives uniqueness. Slater's condition gives KKT necessity, and convexity gives KKT sufficiency. Finally, if \(\gamma(\uk)>\kappa\), then \(\alpha\uk\), with \(\alpha=\kappa/\gamma(\uk)\in(0,1)\), remains feasible and has smaller barrier value along the ray from the origin. This contradiction proves \(\gamma(\uk)=\kappa\). KKT multipliers need not be unique when active examples are linearly dependent, but the primal minimizer is unique.

\subsection{Endpoint Geometry}

Let
\begin{equation}
v_2=\arg\min_v\frac12\|v\|_2^2
\quad\mathrm{s.t.}\quad
z_i^\top v\ge1\ \ \forall i.
\label{eq:l2-separator}
\end{equation}
Also let \(U_\infty=\arg\max_{\|u\|_\infty\le1}\gamma(u)\).

\propheading{Static endpoints}\label{prop:static-endpoints}
As \(\kappa\downarrow0\),
\begin{equation}
\uk/\kappa\to v_2
\label{eq:small-endpoint}
\end{equation}
As \(\kappa\uparrow\ginf\),
\begin{equation}
\operatorname{dist}(\uk,U_\infty)\to0.
\label{eq:set-valued-endpoint}
\end{equation}
If \(U_\infty=\{u_\infty\}\), then
\begin{equation}
\uk\to u_\infty .
\label{eq:large-endpoint}
\end{equation}

\noindent\textit{Proof idea.}
For the small-rate endpoint, set \(u=\kappa v\). The barrier expansion gives
\begin{equation}
B(\kappa v)
=\frac{\kappa^2}{2}\|v\|_2^2
+O(\kappa^3\|v\|^3),
\label{eq:barrier-small-expansion}
\end{equation}
uniformly on bounded \(v\)-sets. Comparing the optimizer with \(\kappa v_2\) bounds \(\uk/\kappa\), and every subsequential limit solves the Euclidean hard-margin problem \eqref{eq:l2-separator}. For the upper endpoint, compactness of the closed cube gives convergent subsequences as \(\kappa\uparrow\ginf\). Passing the inequalities \(Z\uk\ge\kappa\one\) to the limit places every cluster point in \(U_\infty\), which gives set-valued convergence. A singleton \(U_\infty\) gives the pointwise limit in \eqref{eq:large-endpoint}. Without uniqueness, the correct statement is \eqref{eq:set-valued-endpoint}; the data do not specify a single named \(u_\infty\) without an additional tie-breaking rule. The critical case \(\kappa=\ginf\) remains excluded from the open-domain path.

\section{Dual Mirror Ascent Gives the Path}

The dynamic theorem avoids active-set asymptotics by changing variables. Define the scaled potential \(\Phi(u)=\eps_0B(u)\). Its conjugate is
\begin{equation}
\begin{aligned}
\Phi^\star(q)
&=
\sum_{j=1}^d
\left[
|q_j|-\eps_0\log\left(1+\frac{|q_j|}{\eps_0}\right)
\right],\\
\nabla\Phi^\star(q)&=\frac{q}{|q|+\eps_0}.
\end{aligned}
\label{eq:conjugate}
\end{equation}
The equality holds coordinatewise, including value zero when \(q_j=0\). The corresponding concave dual objective is
\begin{equation}
D(\lambda)=\kappa\one^\top\lambda-\Phi^\star(Z^\top\lambda),
\qquad
\lambda\ge0 .
\label{eq:dual-objective}
\end{equation}
Strong duality holds for the static program with \(\Phi\). Dual multipliers may be nonunique, but all dual optimizers have the same image
\begin{equation}
q_\kappa=Z^\top\lambda^\star
=\nabla\Phi(\uk)
=\eps_0\frac{\uk}{1-|\uk|}.
\label{eq:qkappa}
\end{equation}

\theoremheading{Discrete smoothed-sign convergence}\label{thm:dynamic}
Assume \(0<\kappa<\ginf\), the weighted exponential loss \eqref{eq:weighted-exp-loss}, the schedule \eqref{eq:time-schedule}, and the stepsize conditions \eqref{eq:eta-g-condition} with \(\eta_t>0\), \(S_t\to\infty\), and \(\sum_t\eta_t^2<\infty\). Then the memoryless smoothed-sign iterates \eqref{eq:softsign} satisfy
\begin{equation}
\frac{w_t}{S_t}\to\uk,\qquad
e^{\kappa S_t}(-\nabla L(w_t))\to q_\kappa,\qquad
d_t\to\uk,
\label{eq:dynamic-limits}
\end{equation}
where \(d_t=(-\nabla L(w_t))/(|\nabla L(w_t)|+\eps_t)\). Moreover, for finite constants \(C_\Phi,C_D\),
\begin{equation}
\left\|\frac{w_t}{S_t}-\uk\right\|_2
\le
\frac{\|w_0\|_2}{S_t}
+
\sqrt{\frac{C_\Phi C_D}{S_t}} .
\label{eq:finite-time-bound}
\end{equation}
Consequently \(\gamma(w_t/S_t)\to\kappa\), with a finite-time margin bound obtained by multiplying \eqref{eq:finite-time-bound} by \(\max_i\|z_i\|_2\).

\noindent\textit{Principal proof chain.}
Introduce the exact dual variables
\begin{equation}
\lambda_{i,t}=a_i\exp(\kappa S_t-z_i^\top w_t).
\label{eq:lambda-def}
\end{equation}
Then
\begin{equation}
e^{\kappa S_t}(-\nabla L(w_t))=Z^\top\lambda_t,
\qquad
d_t=\nabla\Phi^\star(Z^\top\lambda_t),
\label{eq:exact-identities}
\end{equation}
and the primal update is equivalent to
\begin{equation}
\lambda_{t+1}
=
\lambda_t\odot\exp(\eta_t\nabla D(\lambda_t)).
\label{eq:mirror-update}
\end{equation}
Thus the dynamics is entropic mirror ascent on \eqref{eq:dual-objective}.

Fix any dual optimizer \(\lambda^\star\). Let
\begin{equation}
\begin{aligned}
V_t
&=\sum_i
\left[
\lambda_i^\star\log\frac{\lambda_i^\star}{\lambda_{i,t}}
-\lambda_i^\star+\lambda_{i,t}
\right],\\
\Delta_t&=D^\star-D(\lambda_t).
\end{aligned}
\label{eq:kl-gap}
\end{equation}
The update \eqref{eq:mirror-update} gives the exact one-step identity
\begin{equation}
\begin{aligned}
V_{t+1}-V_t
&=\eta_t\langle\lambda_t-\lambda^\star,\nabla D(\lambda_t)\rangle \\
&\quad+
\sum_i\lambda_{i,t}\!\left[
e^{\eta_t\nabla_iD(\lambda_t)}
-1-\eta_t\nabla_iD(\lambda_t)
\right].
\end{aligned}
\label{eq:kl-recursion}
\end{equation}
Concavity gives the first term at most \(-\eta_t\Delta_t\). Since \(\|\nabla D(\lambda)\|_\infty\le G\), \(\eta_tG\le1\) and \(e^x-1-x\le(e/2)x^2\) imply
\begin{equation}
V_{t+1}-V_t
\le
-\eta_t\Delta_t
+
\frac e2\eta_t^2G^2\|\lambda_t\|_1.
\label{eq:kl-ineq}
\end{equation}
The scalar inequality
\begin{equation}
\|\lambda\|_1
\le
2D_h(\lambda^\star,\lambda)
+2\log2\,\|\lambda^\star\|_1
\label{eq:mass-control}
\end{equation}
controls the noncompact positive orthant, including zero coordinates of \(\lambda^\star\). Since \(\sum_t\eta_t^2<\infty\), \eqref{eq:kl-ineq} and \eqref{eq:mass-control} imply \(\sup_tV_t<\infty\), \(\sup_t\|\lambda_t\|_1<\infty\), and
\begin{equation}
\sum_{s<t}\eta_s\Delta_s\le C_D .
\label{eq:summed-gap}
\end{equation}

The dual gap controls the unique primal direction. Using KKT and Fenchel equality,
\begin{equation}
\Delta_t
=
D_{\Phi^\star}(Z^\top\lambda_t,q_\kappa)
+
\lambda_t^\top(Z\uk-\kappa\one).
\label{eq:gap-decomposition}
\end{equation}
The feasibility term is nonnegative. Bounded dual mass places \(Z^\top\lambda_t\) in a compact box where \(\Phi^\star\) is uniformly strongly convex, while \(\nabla\Phi^\star\) is globally \(1/\eps_0\)-Lipschitz. Hence
\begin{equation}
\|d_t-\uk\|_2^2\le C_\Phi\Delta_t .
\label{eq:direction-gap}
\end{equation}
Finally,
\begin{equation}
w_t=w_0+\sum_{s<t}\eta_sd_s.
\label{eq:primal-average}
\end{equation}
Jensen's inequality, \eqref{eq:summed-gap}, and \eqref{eq:direction-gap} give \eqref{eq:finite-time-bound}. A spike-exclusion lemma using
\(|\Delta_{t+1}-\Delta_t|\le C\eta_t\) upgrades the averaged control to \(\Delta_t\to0\), yielding the last two limits in \eqref{eq:dynamic-limits}. Complete constants and the spike lemma are in the appendix.

\noindent\textbf{Rate diagram scope.}
Theorem~\ref{thm:dynamic} proves the interior regime \(0<\kappa<\ginf\) for weighted exponential loss. The endpoint regimes \(\kappa=0\) and \(\kappa>\ginf\), logistic-loss perturbations, and the critical case \(\kappa=\ginf\) are not included in the theorem. Experiments below keep those diagnostics separate.

\section{Adam as a Boundary Case}

\noindent\textbf{Perturbation lemma.}
The deterministic lemma behind this diagnostic is simple. If
\begin{equation}
x_{t+1}=x_t+\eta_t d_t,
\qquad
y_{t+1}=y_t+\eta_t(d_t+e_t),
\label{eq:perturb-recursions}
\end{equation}
with \(x_0=y_0\), and
\begin{equation}
\frac1{S_t}\sum_{s<t}\eta_s\|e_s\|\to0,
\label{eq:perturb-condition}
\end{equation}
then
\begin{equation}
\frac{\|x_t-y_t\|}{S_t}\to0.
\label{eq:perturb-conclusion}
\end{equation}
Indeed, the triangle inequality gives \(\|x_t-y_t\|\le\sum_{s<t}\eta_s\|e_s\|\). This makes the Adam-transfer residual scientifically relevant: it is the empirical version of \eqref{eq:perturb-condition}. However, E3 did not satisfy the required small averaged residual condition under the prespecified runs. Adam therefore remains a documented empirical boundary case, not part of the main claim or title.

\noindent\textbf{E3 transfer diagnostic.}
The broad Adam-transfer target required
\begin{equation}
\frac{1}{S_t}\sum_{s<t}\eta_s
\left\|
\frac{-m_s}{\sqrt{v_s}+\eps_s}
-
\frac{-\nabla L(w_s)}{|\nabla L(w_s)|+\eps_s}
\right\|\to0 .
\label{eq:adam-transfer}
\end{equation}
If \eqref{eq:adam-transfer} held under the same assumptions as Theorem~\ref{thm:dynamic}, the perturbation lemma would transfer the smoothed-sign limit. The prespecified E3 experiments did not support \eqref{eq:adam-transfer} under the tested schedules, so no Adam theorem is claimed.

\section{Fixed-\(\epsilon\) Crossover}

\noindent\textbf{Definition and numerical target.}
For fixed \(\eps>0\), define the primary transition
\begin{equation}
\tau_\eps(c)=\min\{t:\|-\nabla L(w_t)\|_\infty\le c\eps\},
\quad c\in\{0.5,1,2\}.
\label{eq:tau}
\end{equation}
The target law is
\begin{equation}
S_{\tau_\eps(c)}/\log(1/\eps)\to1/\ginf .
\label{eq:crossover}
\end{equation}
Equation~\eqref{eq:crossover} is numerically supported; the upper and lower proof bounds for the fixed-\(\epsilon\) dynamics remain open.

For clarity, \(\tau_\eps(c)\) in \eqref{eq:tau} is a gradient threshold for the fixed-\(\epsilon\) memoryless smoothed-sign dynamics. It is not an Adam second-moment threshold and does not involve \(\sqrt{v_t}\).

\noindent\textbf{Conditional crossover implication.}
Suppose that, uniformly up to the transition window,
\begin{equation}
c_0e^{-\ginf S_t}
\le
\|-\nabla L(w_t)\|_\infty
\le
Ce^{-\ginf S_t}.
\label{eq:crossover-tail-bounds}
\end{equation}
Then the threshold relation \(\|-\nabla L(w_t)\|_\infty\le c\eps\) occurs only after the lower bound can fall below \(c\eps\), and occurs once the upper bound falls below \(c\eps\). Thus, up to the \(S_t\)-mesh,
\begin{equation}
\frac1{\ginf}\log\frac1\eps+\frac1{\ginf}\log\frac{c_0}{c}
\le
S_{\tau_\eps(c)}
\le
\frac1{\ginf}\log\frac1\eps+\frac1{\ginf}\log\frac{C}{c}.
\label{eq:crossover-bracket}
\end{equation}
For fixed \(c\), this implies
\begin{equation}
S_{\tau_\eps(c)}
=\frac1{\ginf}\log\frac1\eps+O(1),
\qquad
\frac{S_{\tau_\eps(c)}}{\log(1/\eps)}
\to
\frac1{\ginf}.
\label{eq:crossover-conditional-law}
\end{equation}
This is a conditional implication, not a proved tail bound.

\noindent\textbf{Proof gap.}
The unresolved obstacles are changes in the leading coordinate and lack of uniform tail bounds up to the transition time.

\section{Experiments}

\subsection{Protocol and Metrics}

The primary evidence uses only saved outputs generated from deterministic seeds, float64 arithmetic, numerical separability checks, positive exponential-loss weights, configuration hashes, solver-status logging, and KKT residual reporting. Figures are generated from saved data with metadata sidecars; plotted values are not edited by hand. The illustrative two-dimensional dataset in Figure~\ref{fig:concept} is excluded from all aggregate statistics.

The theorem-compatible validation suite uses three synthetic families: isotropic random separable data, controlled-support-vector data, and correlated ill-conditioned data. The aggregate grid has \(n=4d\), \(d\in\{10,25,50,100\}\), twenty unfiltered seeds per setting, weights \(a_i=1\) and bounded log-uniform positive weights, and
\[
\kappa/\ginf\in\{0.10,0.25,0.50,0.75,0.90\}.
\]
The primary schedule is \(\eta_t=\eta_0(t+t_0)^{-0.75}\) with \(\eta_0=0.5/G\), so \(\eta_tG\le0.5\), \(S_t\to\infty\), and \(\sum_t\eta_t^2<\infty\). Secondary theorem-compatible powers \(0.60\) and \(0.90\) are used only for schedule checks. The numerical barrier solve uses the open-domain proxy
\begin{equation}
|u_j|\le1-\delta,
\qquad
\delta=10^{-7}.
\label{eq:numerical-barrier-delta}
\end{equation}
Convex-program failures are retained in the output and separated from dynamic path errors; only successful barrier solves are used as primary evidence for Theorem~\ref{thm:dynamic}.

The main theorem diagnostics are evaluated at multiple cumulative-time checkpoints. We report
\begin{equation}
E_{\mathrm{abs}}(t)=
\left\|
\frac{w_t}{S_t}-\uk
\right\|_2 ,
\qquad
E_d(t)=\|d_t-\uk\|_2,
\label{eq:p2-main-metrics-a}
\end{equation}
along with
\begin{equation}
\begin{aligned}
E_q(t)&=
\left\|
e^{\kappa S_t}(-\nabla L(w_t))-q_\kappa
\right\|_2,\\
E_\gamma(t)&=
\left|
\gamma\!\left(\frac{w_t}{S_t}\right)-\kappa
\right|.
\end{aligned}
\label{eq:p2-main-metrics-b}
\end{equation}
The relative and angular errors are reported in the appendix. The exact dual-transformation residuals are
\begin{equation}
\begin{aligned}
r_q(t)&=
\left\|e^{\kappa S_t}(-\nabla L(w_t))-Z^\top\lambda_t\right\|_2,\\
r_\lambda(t)&=
\left\|\log\lambda_{t+1}-\log\lambda_t-\eta_t\nabla D(\lambda_t)\right\|_\infty .
\end{aligned}
\label{eq:identity-residual-metrics}
\end{equation}
For the fixed-\(\epsilon\) crossover, each threshold family fits
\begin{equation}
S_{\tau_\eps(c)}=b_0+b_1\log(1/\eps),
\qquad
\rho_{\mathrm{cross}}=\frac{b_1}{1/\ginf}=b_1\ginf .
\label{eq:crossover-slope-ratio}
\end{equation}
The reported crossover slope ratio \(\rho_{\mathrm{cross}}\) should be close to one under the empirical scaling law.

\subsection{Direct Validation of Theorem 1}

\begin{figure*}[t]
\centering
\includegraphics[width=\textwidth]{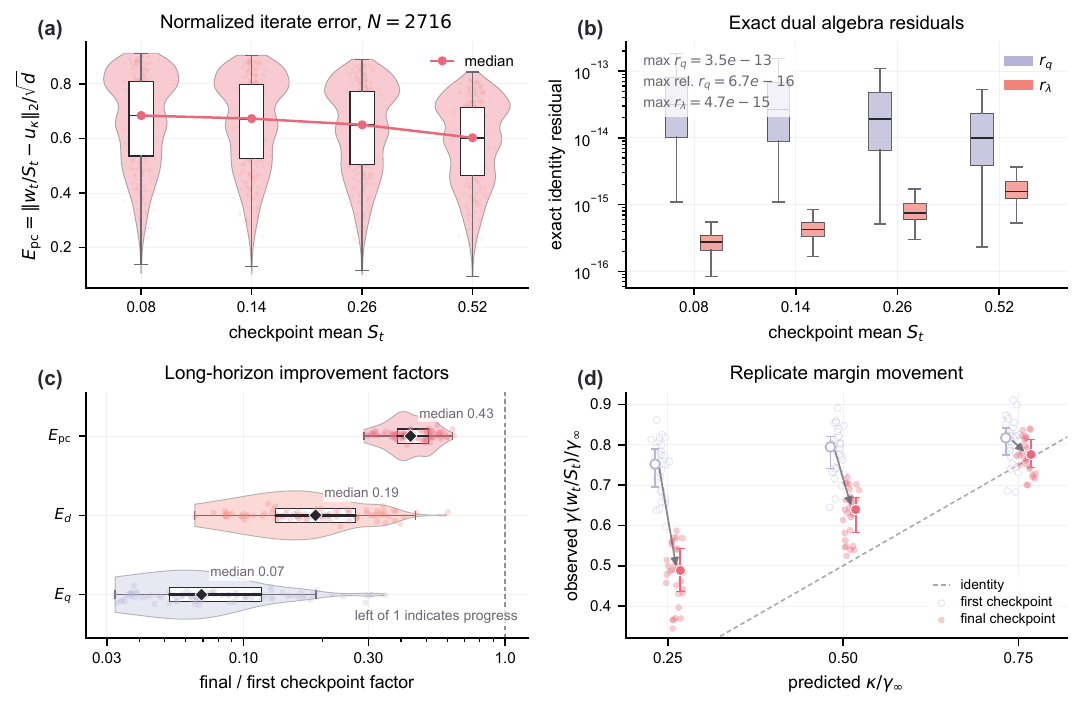}
\caption{Theorem validation dashboard. All panels use theorem-compatible weighted exponential loss, \(0<\kappa<\ginf\), square-summable schedules, and \(\eta_tG\le0.5\). Panels (a,b) use the primary aggregate suite with \(2716\) successful barrier targets: panel (a) displays the distribution of \(E_{\rm pc}=E_{\rm abs}/\sqrt d\) rather than a selected seed, and panel (b) checks exact dual algebra residuals \(r_q\) and \(r_\lambda\) at floating-point scale. Panels (c,d) use the saved \(89\)-target long-horizon showcase without additional runs: panel (c) reports final-over-first improvement-factor distributions for \(E_{\rm pc}\), \(E_d\), and \(E_q\), where values below \(1\) indicate finite-window progress, and panel (d) overlays replicate-level first/final normalized margins with median movement arrows against the predicted \(\kappa/\ginf\) identity line. No sharp empirical rate is claimed; certified-envelope and shorter-horizon margin diagnostics are reported in the appendix.}
\label{fig:theorem-validation}
\end{figure*}

The primary run completed all \(3200\) dynamic trajectories and obtained \(2716\) successful barrier targets. Panels (a,b) therefore emphasize breadth: the aggregate distribution is shown across successful theorem-compatible targets rather than through a selected seed. The remaining \(484\) primary records are retained as numerical barrier-boundary cases and are excluded from the theorem-evidence aggregation. Exact dual residuals remain at floating-point scale.

Panels (c,d) add depth using the separate long-horizon showcase. That run attempted \(90\) theorem-compatible configurations, retained one failed barrier solve, and used the \(89\) successful targets for evidence panels. The saved horizon extends the maximum cumulative time from \(S_t=1.09\) in the primary grid to \(S_t=4.38\), and the final/first diagnostic factors and replicate-level normalized-margin movements continue to point in the theorem-predicted direction over that finite window. The certified finite-time constants remain conservative and are treated as an appendix envelope diagnostic. We do not claim that the empirical rate is \(S_t^{-1/2}\) or that the proof constants are sharp.

\subsection{Mechanism Diagnostics}

\begin{figure*}[t]
\centering
\includegraphics[width=.88\textwidth]{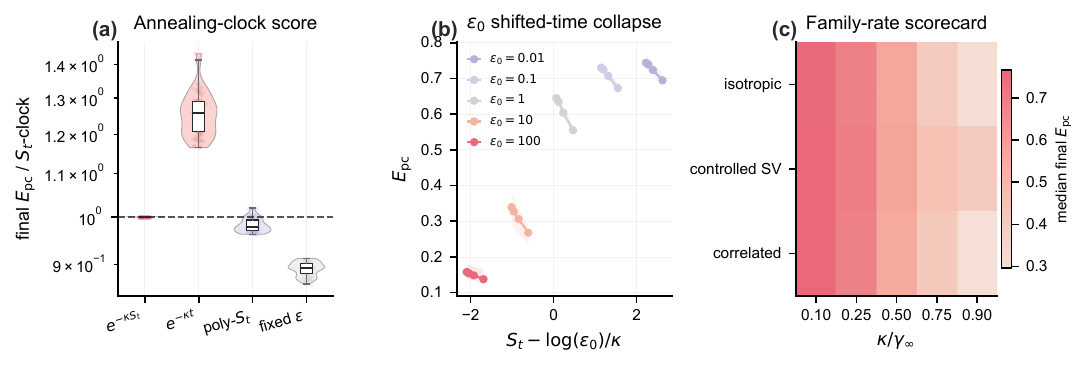}
\caption{Mechanism scorecard. Panel (a) compares final finite-window error ratios against the proposed exponential \(S_t\)-clock reference. Panel (b) shows shifted-time transients across \(\epsilon_0\) values, indicating partial alignment rather than an exact time translation. Panel (c) summarizes median final \(E_{\rm pc}\) by data family and rate. These are empirical mechanism diagnostics, not new theorems.}
\label{fig:mechanism}
\end{figure*}

Figure~\ref{fig:mechanism} tests whether the observed path behavior is tied to the annealing clock and target geometry rather than only to post-hoc fitting. The exponential \(S_t\)-schedule is the reference; alternatives are compared by paired finite-window error ratios. The \(\epsilon_0\) panel shows target invariance with finite-window speed differences and partial shifted-time alignment. The family/rate panel is a robustness scorecard. The square-root smoothing-map cross-fit is mixed and remains in the appendix as a boundary diagnostic; no square-root-map theorem is claimed.

\subsection{Fixed-\(\epsilon\) Crossover}

\begin{figure*}[t]
\centering
\includegraphics[width=.88\textwidth]{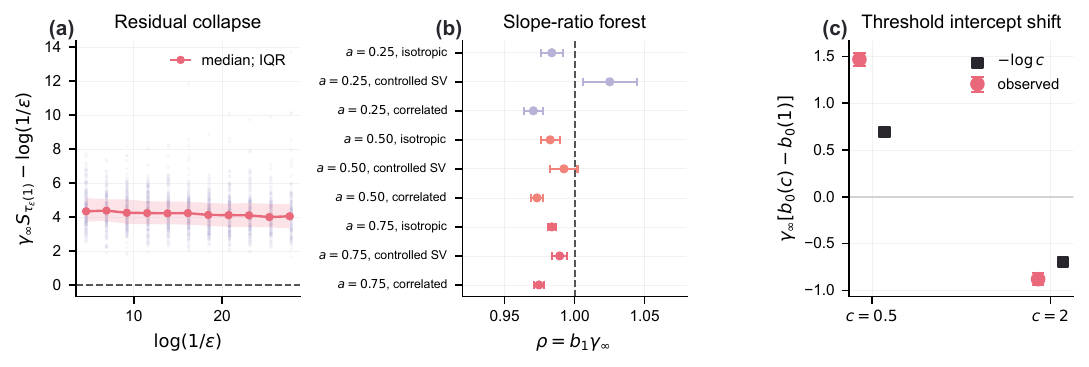}
\caption{Fixed-\(\epsilon\) crossover scaling for the memoryless proxy, with \(\tau_\eps(c)=\min\{t:\|-\nabla L(w_t)\|_\infty\le c\epsilon\}\). Panel (a) shows residual collapse for the empirical law \(S_{\tau_\epsilon}=\ginf^{-1}\log(1/\epsilon)+O(1)\). Panel (b) reports \(\rho=b_1\ginf\) by learning-rate power and family, and panel (c) compares normalized threshold-induced intercept shifts with \(-\log c\). This remains empirical and conditional; the two-sided tail bound needed for a proof is open.}
\label{fig:crossover}
\end{figure*}

The crossover suite uses \(d\in\{10,25,50\}\), all three synthetic families, twenty seeds, \(\epsilon\in\{10^{-2},10^{-3},\ldots,10^{-12}\}\), \(c\in\{0.5,1,2\}\), and powers \(a\in\{0.25,0.50,0.75\}\). Thresholds were reached in \(99.23\%\) of saved transition records. The residual panel supports \(S_{\tau_\epsilon}=\ginf^{-1}\log(1/\epsilon)+O(1)\) empirically. Across all \(\epsilon\) values and \(c=1\), mean slope ratios were \(0.994\), \(0.987\), and \(0.985\) for \(a=0.25,0.50,0.75\), with subgroup confidence intervals centered near one. Changing \(c\) primarily changed the intercept, and the normalized shifts move in the direction predicted by \(-\log c\). Sensitivity analyses that remove the largest or smallest \(\epsilon\) are kept in the appendix and are not used to tune the reported range.

\subsection{Robustness and Boundary Diagnostics}

Beyond the theorem-compatible exponential-loss runs, the remaining sweeps delineate scope rather than enlarge Theorem~\ref{thm:dynamic}. Logistic-loss runs test whether the observed exponential-tail behavior is numerically stable under another common classification tail, but they are not covered by the theorem. Schedule, \(\epsilon_0\), data-family, dimension, support-vector, rescaling, and boundary-pathology sweeps are reported in the appendix as finite-horizon diagnostics. They found no numerical counterexample within the prespecified grid, but they are not used to claim convergence outside the weighted-exponential, memoryless smoothed-sign setting.

The adaptive-method diagnostics serve a different purpose. The perturbation lemma shows that Adam or RMSProp would inherit the smoothed-sign limit only if the averaged direction residual were small on the \(S_t\)-scale. The prespecified E3 comparison measured both path error to the same \(\uk\) and the weighted perturbation residual \(S_t^{-1}\sum_{s<t}\eta_s\Rdyn(s)\). Mean final path error was \(0.0886\) for smoothed-sign descent, \(0.8432\) for RMSProp, and \(0.8443\) for Adam; mean weighted transfer residuals were \(3.8404\) for RMSProp and \(3.8668\) for Adam. These results do not support broad Adam transfer under the tested grid, but they do not prove an impossibility theorem for every adaptive variant.

The fixed-\(\epsilon\) crossover should be read in the same boundary-diagnostic spirit. Figure~\ref{fig:crossover} supports the empirical law \(S_{\tau_\epsilon}=\ginf^{-1}\log(1/\epsilon)+O(1)\) for the memoryless proxy, but the proof would require two-sided gradient-tail control up to the threshold window. The appendix records these missing obligations explicitly. Thus the crossover evidence clarifies how long fixed-stability transients can look sign-like; it does not broaden the annealed-stability theorem.

\section{Limitations and Conclusions}

The theorem isolates \(0<\kappa<\ginf\) for weighted exponential loss and memoryless stability-annealed smoothed-sign descent. The endpoint regimes \(\kappa=0\) and \(\kappa>\ginf\), the critical case \(\kappa=\ginf\), logistic perturbation theory, fixed-\(\epsilon\) tail bounds, and full Adam transfer require separate arguments. Numerically, the horizon is finite, endpoint barrier solves can fail, and aggregate experiments are synthetic separable linear-classification instances matching the theorem setting.

Within this scope, the result gives a concrete answer to the stability-annealing question. The exponential decay rate of the stability constant becomes a margin constraint, and the coordinatewise smoothing map induces the Burg-type barrier that selects the point on that margin slice. The static program is well posed, and the dynamic theorem follows by rewriting the iterates as entropic mirror ascent with a KL recursion. This separates the proved selector from adjacent empirical phenomena: logistic tails, fixed-\(\epsilon\) crossover, and adaptive-method behavior remain informative diagnostics rather than theorem claims.

The cumulative-time clock is essential: theorem and experiments compare the gradient tail and stability constant on the same exponential \(S_t\)-scale, where schedules similar in iteration time can differ over finite \(S_t\)-windows. The experiments target exact dual algebra, the rate-indexed path, and boundary behavior; they are not large-scale benchmarks.

\clearpage

\clearpage
\bibliography{references}

\clearpage
\appendix

\section{A. Complete Assumptions and Notation}

The main paper gives a shorter assumption block for readability. This appendix repeats the full ledger so that the supplement can be read independently.

\paragraph{Proof-tool references.}
The static proof uses standard KKT, Slater, and Fenchel-duality facts for convex programs. The dynamic proof uses standard Bregman, mirror-descent, exponentiated-gradient, and multiplicative-weights notation.

\paragraph{Signed data and margins.}
The signed examples are fixed vectors \(z_i=y_i x_i\in\R^d\), collected as rows of \(Z\in\R^{n\times d}\). The raw margin and \(\ell_\infty\) max-margin value are
\begin{equation}
\gamma(u)=\min_{i\in[n]} z_i^\top u,
\qquad
\ginf=\max_{\|u\|_\infty\le1}\gamma(u).
\label{eq:app-margin}
\end{equation}
The static barrier program assumes separability in the form \(\ginf>0\) and uses rates \(0<\kappa<\ginf\). The critical case \(\kappa=\ginf\) is excluded unless stated separately.

\paragraph{Dynamic theorem assumptions.}
The exact dynamic theorem is for weighted exponential loss
\begin{equation}
L(w)=\sum_{i=1}^n a_i e^{-z_i^\top w},
\qquad a_i>0.
\label{eq:app-weighted-exp}
\end{equation}
The cumulative learning-rate time and annealed stability schedule are
\begin{equation}
S_t=\sum_{s=0}^{t-1}\eta_s,
\qquad
\eps_t=\eps_0\exp(-\kappa S_t).
\label{eq:app-time}
\end{equation}
It assumes \(\eta_t>0\), \(S_t\to\infty\), and \(\sum_t\eta_t^2<\infty\). With
\[
G=\max_i(\kappa+\|z_i\|_1),
\]
it also assumes \(\eta_tG\le1\) for every \(t\). A finite-prefix relaxation is not used in the submitted theorem. Logistic loss is an empirical robustness check only; no logistic perturbation theorem is claimed.

\paragraph{Memoryless smoothed-sign theorem.}
The central dynamics is
\begin{equation}
w_{t+1}=w_t+\eta_t
\frac{-\nabla L(w_t)}{|\nabla L(w_t)|+\eps_t},
\label{eq:app-soft}
\end{equation}
where absolute values and divisions are coordinatewise.

\paragraph{Static barrier assumptions.}
The barrier is
\begin{equation}
B(u)=\sum_{j=1}^d[-|u_j|-\log(1-|u_j|)],
\qquad u\in(-1,1)^d .
\label{eq:app-barrier}
\end{equation}
The theoretical domain is open. Numerical solves use the prespecified closed approximation \(|u_j|\le1-\delta\), with \(\delta=10^{-7}\), and record solver status and KKT residuals.

\paragraph{Non-assumptions of the dynamic theorem.}
The proof does not assume bounded normalized iterates, active-set stability, active-prefactor convergence, absence of coordinate cancellation, subsequence identification, strict complementarity, unique dual multipliers, or generic-position data. The dual mirror-ascent proof below replaces those heuristic requirements.

\paragraph{Adam and RMSProp scope.}
Full-batch Adam and RMSProp are used as prespecified diagnostics. No full-Adam theorem is claimed. The broad transfer target from the memoryless proxy to Adam is not supported by E3 under the tested schedules. Future Adam claims would need a narrower statement and a new proof.

\paragraph{Fixed-\(\epsilon\) crossover scope.}
For fixed \(\eps>0\), the primary transition definition is
\begin{equation}
\tau_\eps(c)=\min\{t:\|-\nabla L(w_t)\|_\infty\le c\eps\},
\qquad c\in\{0.5,1,2\}.
\label{eq:app-tau}
\end{equation}
This is a gradient-threshold definition for the fixed-\(\epsilon\) memoryless smoothed-sign proxy. It is not an Adam second-moment threshold and does not involve \(\sqrt{v_t}\).

\section{B. Static Barrier Geometry and Endpoint Proofs}

\paragraph{Scalar barrier facts.}
Let \(b(r)=-r-\log(1-r)\) for \(r\in[0,1)\). Then
\begin{equation}
b'(r)=\frac{r}{1-r},\qquad
b''(r)=\frac{1}{(1-r)^2}.
\label{eq:app-scalar-derivatives}
\end{equation}
Thus \(b(0)=b'(0)=0\), \(b(r)\ge0\), \(b\) is increasing, and \(b\) is strictly convex away from the origin. Since \(b''(r)\ge1\), the coordinate potential \(u\mapsto b(|u|)\) is \(1\)-strongly convex on \((-1,1)\). The one-sided first derivatives at zero agree, so \(B\) is differentiable on the open cube and
\begin{equation}
[\nabla B(u)]_j=\frac{u_j}{1-|u_j|},
\label{eq:app-grad}
\end{equation}
with value zero at \(u_j=0\). Consequently \(B\) is convex and \(1\)-strongly convex on \((-1,1)^d\) in the Euclidean sense.

\subsection{Proof of Proposition 1 (Well-posedness and KKT)}

The program is
\begin{equation}
\min_{u\in(-1,1)^d} B(u)
\quad\mathrm{s.t.}\quad
Zu\ge\kappa\one .
\label{eq:app-program}
\end{equation}
Let \(u_\infty\) be any maximizer of \(\gamma\) over the closed cube. Because \(0<\kappa<\ginf\), the scaled point
\begin{equation}
\tilde u=\frac12\left(1+\frac{\kappa}{\ginf}\right)u_\infty
\label{eq:app-strict-feasible}
\end{equation}
satisfies \(\|\tilde u\|_\infty<1\) and
\[
Z\tilde u\ge \frac{\ginf+\kappa}{2}\one>\kappa\one .
\]
Hence Slater's condition holds for the margin constraints.

To handle the open domain cleanly, define the extended-value closure
\[
\bar B(u)=
\begin{cases}
B(u), & \|u\|_\infty<1,\\
+\infty, & \|u\|_\infty=1,
\end{cases}
\]
and minimize \(\bar B\) over the compact convex set
\[
K_\kappa=\{u:\|u\|_\infty\le1,\ Zu\ge\kappa\one\}.
\]
The strictly feasible point \(\tilde u\) gives a finite objective value. Boundary points have infinite value because at least one term \(-\log(1-|u_j|)\) diverges, so every minimizer of the closed extended-value problem lies in the open cube. It is therefore a minimizer of the original open-domain program. Strict convexity of \(B\) on the convex feasible set gives uniqueness; call the unique minimizer \(\uk\).

The margin of \(\uk\) is exactly \(\kappa\). If instead \(\gamma(\uk)>\kappa\), choose \(\alpha=\kappa/\gamma(\uk)\in(0,1)\). Then \(\alpha\uk\) remains feasible because \(\gamma(\alpha\uk)=\kappa\). Since \(\uk\neq0\) and every nonzero coordinate has \(b(|\alpha u_j|)<b(|u_j|)\), we get \(B(\alpha\uk)<B(\uk)\), contradicting optimality. This radial-scaling argument proves that at least one margin constraint is active; it does not assert that every constraint is active.

Writing the inequality as \(\kappa\one-Zu\le0\), the Lagrangian is
\[
\mathcal{L}(u,\lambda)=B(u)+\lambda^\top(\kappa\one-Zu),
\qquad \lambda\ge0 .
\]
Slater's condition and differentiability of \(B\) at the optimizer imply KKT necessity:
\begin{align}
Z\uk&\ge\kappa\one,\qquad \lambda\ge0,\notag\\
\lambda_i(z_i^\top\uk-\kappa)&=0,\qquad
\nabla B(\uk)=Z^\top\lambda .
\label{eq:app-kkt}
\end{align}
Conversely, if a feasible \(u\) and multiplier \(\lambda\ge0\) satisfy \eqref{eq:app-kkt}, then for any feasible \(v\),
\begin{align*}
B(v)&\ge B(u)+\langle\nabla B(u),v-u\rangle\\
&=B(u)+\lambda^\top Z(v-u)\ge B(u),
\end{align*}
where complementarity gives \(\lambda^\top(Zu-\kappa\one)=0\) and feasibility gives \(\lambda^\top(Zv-\kappa\one)\ge0\). Thus the KKT system is sufficient, and strict convexity gives the unique primal solution.

\subsection{Proof of Proposition 2 (Static endpoints)}

\paragraph{Small-rate endpoint.}
Let
\begin{equation}
v_2=\arg\min_v\frac12\|v\|_2^2
\quad\mathrm{s.t.}\quad Zv\ge\one .
\label{eq:app-v2}
\end{equation}
The minimizer is unique because the objective is strictly convex and the feasible set is closed and convex. With \(u=\kappa v\), the constraints \(Zu\ge\kappa\one\) become \(Zv\ge\one\). Near the origin,
\begin{equation}
B(\kappa v)
=\frac{\kappa^2}{2}\|v\|_2^2
+\frac{\kappa^3}{3}\sum_j |v_j|^3
+O(\kappa^4\|v\|^4),
\label{eq:app-small-expansion}
\end{equation}
uniformly on bounded \(v\)-sets. Optimality of \(\uk\) against \(\kappa v_2\) gives \(B(\uk)\le B(\kappa v_2)=O(\kappa^2)\). Since \(b''(r)\ge1\) and \(b(0)=b'(0)=0\), \(B(u)\ge\frac12\|u\|_2^2\); hence \(v_\kappa=\uk/\kappa\) is bounded.

Take any convergent subsequence \(v_{\kappa_m}\to\bar v\) with \(\kappa_m\downarrow0\). Feasibility gives \(Z\bar v\ge\one\). For any feasible \(v\), compare \(\uk\) with \(\kappa v\) and divide by \(\kappa^2\). Passing to the limit using \eqref{eq:app-small-expansion} gives
\[
\frac12\|\bar v\|_2^2\le \frac12\|v\|_2^2 .
\]
Thus \(\bar v=v_2\). Every subsequential limit is \(v_2\), so
\begin{equation}
\frac{\uk}{\kappa}\to v_2
\qquad \text{as }\kappa\downarrow0 .
\label{eq:app-small-endpoint}
\end{equation}

\paragraph{Upper endpoint.}
Let
\[
U_\infty=\arg\max_{\|u\|_\infty\le1}\gamma(u).
\]
For any sequence \(\kappa_m\uparrow\ginf\), compactness of the closed cube gives a convergent subsequence \(u_{\kappa_m}\to\bar u\). Since \(Zu_{\kappa_m}\ge\kappa_m\one\), continuity gives \(Z\bar u\ge\ginf\one\), hence \(\gamma(\bar u)\ge\ginf\). By definition of \(\ginf\), \(\gamma(\bar u)\le\ginf\), so \(\bar u\in U_\infty\). Therefore
\begin{equation}
\dist(\uk,U_\infty)\to0
\qquad \text{as }\kappa\uparrow\ginf .
\label{eq:app-upper-endpoint}
\end{equation}
If \(U_\infty=\{u_\infty\}\), then every cluster point equals \(u_\infty\), and \(\uk\to u_\infty\). If \(U_\infty\) is not a singleton, convergence to a preselected point \(u_\infty\) is not valid; only the set-valued statement \eqref{eq:app-upper-endpoint} is claimed.

\section{C. Exact Dual Mirror-Ascent Dynamic Proof}

This section gives the complete discrete proof for \eqref{eq:app-soft} under the theorem assumptions in Appendix A. The proof is for weighted exponential loss \eqref{eq:app-weighted-exp}.
It does not rely on the fixed-\(\epsilon\) crossover proof program in Appendix E.

\paragraph{Fenchel conjugate and dual objective.}
Let \(\Phi(u)=\eps_0B(u)\). Its Fenchel conjugate is
\begin{equation}
\begin{aligned}
\Phi^\star(q)
&=
\sum_j\left[
|q_j|-\eps_0\log\left(1+\frac{|q_j|}{\eps_0}\right)
\right],\\
\nabla\Phi^\star(q)&=\frac{q}{|q|+\eps_0}.
\end{aligned}
\label{eq:app-phistar}
\end{equation}
The Hessian is diagonal with entries
\begin{equation}
\nabla^2\Phi^\star(q)_{jj}
=
\frac{\eps_0}{(\eps_0+|q_j|)^2}.
\label{eq:app-phistar-hessian}
\end{equation}
The dual objective is the concave function
\begin{equation}
D(\lambda)=\kappa\one^\top\lambda-\Phi^\star(Z^\top\lambda),
\qquad
\lambda\ge0.
\label{eq:app-dual}
\end{equation}
Strong duality follows from Slater's condition for the primal program. If \(\lambda^\star\) is any dual optimizer, then
\begin{equation}
Z^\top\lambda^\star
=q_\kappa
=\nabla\Phi(\uk)
=\eps_0\frac{\uk}{1-|\uk|}.
\label{eq:app-qkappa}
\end{equation}
The multiplier \(\lambda^\star\) may be nonunique, but the image \(q_\kappa\) is unique.

\paragraph{Exact dynamic transformation.}
Define
\begin{equation}
\lambda_{i,t}=a_i\exp(\kappa S_t-z_i^\top w_t),
\qquad
q_t=Z^\top\lambda_t.
\label{eq:app-lambda}
\end{equation}
Then
\begin{equation}
e^{\kappa S_t}(-\nabla L(w_t))=q_t,
\qquad
d_t=\nabla\Phi^\star(q_t).
\label{eq:app-identities}
\end{equation}
Moreover,
\begin{equation}
\lambda_{t+1}
=
\lambda_t\odot\exp(\eta_t\nabla D(\lambda_t)).
\label{eq:app-mirror-update}
\end{equation}
This is entropic mirror ascent on the nonnegative orthant for the entropy \(h(\lambda)=\sum_i\lambda_i(\log\lambda_i-1)\).

\paragraph{KL one-step identity and inequality.}
Fix a dual optimizer \(\lambda^\star\), and write
\begin{equation}
\begin{aligned}
V_t
&=D_h(\lambda^\star,\lambda_t)\\
&=
\sum_i
\left[
\lambda_i^\star\log\frac{\lambda_i^\star}{\lambda_{i,t}}
-\lambda_i^\star+\lambda_{i,t}
\right],\\
\Delta_t&=D^\star-D(\lambda_t).
\end{aligned}
\label{eq:app-kl}
\end{equation}
Zero coordinates of \(\lambda^\star\) contribute only \(\lambda_{i,t}\). With \(g_t=\nabla D(\lambda_t)\), \eqref{eq:app-mirror-update} gives
\begin{equation}
\begin{aligned}
V_{t+1}-V_t
&=\eta_t\langle\lambda_t-\lambda^\star,g_t\rangle\\
&\quad+
\sum_i\lambda_{i,t}
\left[
e^{\eta_tg_{i,t}}-1-\eta_tg_{i,t}
\right].
\end{aligned}
\label{eq:app-kl-identity}
\end{equation}
Concavity of \(D\) gives \(\langle\lambda_t-\lambda^\star,g_t\rangle\le-\Delta_t\). Since
\[
\nabla_iD(\lambda)=\kappa-z_i^\top\nabla\Phi^\star(Z^\top\lambda),
\]
and \(\|\nabla\Phi^\star(q)\|_\infty<1\), we have
\[
\|\nabla D(\lambda)\|_\infty\le G.
\]
Using \(\eta_tG\le1\) and \(e^x-1-x\le(e/2)x^2\) for \(|x|\le1\),
\begin{equation}
V_{t+1}-V_t
\le
-\eta_t\Delta_t
+
\frac e2\eta_t^2G^2\|\lambda_t\|_1.
\label{eq:app-kl-ineq}
\end{equation}

\paragraph{Noncompact dual-mass control and summed gap.}
For \(x\ge0\) and \(y>0\),
\[
y\le2[x\log(x/y)-x+y]+2(\log2)x,
\]
where the bracket is \(y\) when \(x=0\). Summing gives
\begin{equation}
\|\lambda\|_1
\le
2D_h(\lambda^\star,\lambda)
+2\log2\,\|\lambda^\star\|_1.
\label{eq:app-mass}
\end{equation}
Let
\[
E_2=\sum_t\eta_t^2,\quad
L_\star=\|\lambda^\star\|_1,\quad
V_0=D_h(\lambda^\star,\lambda_0),
\]
and define
\begin{equation}
\begin{aligned}
\bar V
&=
\left(V_0+eG^2(\log2)L_\star E_2\right)e^{eG^2E_2},\\
\bar M&=2\bar V+2(\log2)L_\star.
\end{aligned}
\label{eq:app-vbar}
\end{equation}
Equations \eqref{eq:app-kl-ineq} and \eqref{eq:app-mass}, together with discrete Gronwall, imply
\begin{equation}
\sup_tV_t\le\bar V,
\qquad
\sup_t\|\lambda_t\|_1\le\bar M.
\label{eq:app-bounds}
\end{equation}
Rearranging \eqref{eq:app-kl-ineq} and summing gives
\begin{equation}
\sum_{s<t}\eta_s\Delta_s
\le
C_D
:=
V_0+\frac e2G^2\bar M E_2.
\label{eq:app-cd}
\end{equation}

\paragraph{Dual gap controls the primal direction.}
KKT conditions and Fenchel equality give
\begin{equation}
\Delta_t
=
D_{\Phi^\star}(q_t,q_\kappa)
+
\lambda_t^\top(Z\uk-\kappa\one).
\label{eq:app-gap-decomp}
\end{equation}
The feasibility term is nonnegative. Let
\[
Z_\infty=\max_i\|z_i\|_\infty,
\qquad
Q=\max\{\|q_\kappa\|_\infty,\bar MZ_\infty\}.
\]
Then \(q_t\) and \(q_\kappa\) lie in \(\|q\|_\infty\le Q\), where \eqref{eq:app-phistar-hessian} gives
\[
\nabla^2\Phi^\star(q)\succeq mI,
\qquad
m=\frac{\eps_0}{(\eps_0+Q)^2}.
\]
Thus
\[
D_{\Phi^\star}(q_t,q_\kappa)
\ge
\frac m2\|q_t-q_\kappa\|_2^2.
\]
The global upper Hessian bound gives
\[
\|\nabla\Phi^\star(q_t)-\nabla\Phi^\star(q_\kappa)\|_2
\le
\eps_0^{-1}\|q_t-q_\kappa\|_2.
\]
Since \(d_t=\nabla\Phi^\star(q_t)\) and \(\uk=\nabla\Phi^\star(q_\kappa)\),
\begin{equation}
\|d_t-\uk\|_2^2
\le
C_\Phi\Delta_t,
\qquad
C_\Phi=\frac{2}{\eps_0^2m}
=\frac{2(\eps_0+Q)^2}{\eps_0^3}.
\label{eq:app-cphi}
\end{equation}

\paragraph{Jensen normalized-iterate bound.}
The primal recursion is
\[
w_t=w_0+\sum_{s<t}\eta_sd_s.
\]
Jensen's inequality, \eqref{eq:app-cd}, and \eqref{eq:app-cphi} give
\begin{equation}
\left\|
\frac{w_t-w_0}{S_t}-\uk
\right\|_2^2
\le
\frac{C_\Phi C_D}{S_t}.
\label{eq:app-average-bound}
\end{equation}
Therefore
\begin{equation}
\left\|
\frac{w_t}{S_t}-\uk
\right\|_2
\le
\frac{\|w_0\|_2}{S_t}
+
\sqrt{\frac{C_\Phi C_D}{S_t}}.
\label{eq:app-main-bound}
\end{equation}

\paragraph{Spike exclusion, last iterate, and rescaled gradient.}
The dual gap has bounded variation. Since \(\|\nabla D\|_\infty\le G\) and \(|e^x-1|\le e|x|\) for \(|x|\le1\),
\[
|\Delta_{t+1}-\Delta_t|
\le eG^2\bar M\,\eta_t.
\]
We use the elementary spike-exclusion lemma: if \(a_t\ge0\), \(\sum_t\eta_t=\infty\), \(\eta_t\to0\), \(\sum_t\eta_ta_t<\infty\), and \(|a_{t+1}-a_t|\le C\eta_t\), then \(a_t\to0\). Otherwise infinitely many spikes \(a_t\ge2\epsilon\) produce disjoint intervals of cumulative stepsize at least \(\epsilon/(2C)\) on which \(a_s\ge\epsilon\), contradicting weighted summability. Applying the lemma to \(a_t=\Delta_t\) gives \(\Delta_t\to0\). Hence \(q_t\to q_\kappa\), \(d_t\to\uk\), and by \eqref{eq:app-identities}
\[
e^{\kappa S_t}(-\nabla L(w_t))\to q_\kappa.
\]

\paragraph{Margin and loss tails.}
With \(Z_2=\max_i\|z_i\|_2\), \(\gamma\) is \(Z_2\)-Lipschitz, so \eqref{eq:app-main-bound} implies
\[
\left|\gamma(w_t/S_t)-\kappa\right|
\le
Z_2\left[
\frac{\|w_0\|_2}{S_t}
+
\sqrt{\frac{C_\Phi C_D}{S_t}}
\right].
\]
Also
\[
L(w_t)=e^{-\kappa S_t}\|\lambda_t\|_1.
\]
The upper bound follows from \(\|\lambda_t\|_1\le\bar M\). For the lower bound, any dual optimizer is nonzero because \(D^\star=\Phi(\uk)>0\). Choose \(i_0\) with \(\lambda_{i_0}^\star>0\). Since \(V_t\le\bar V\),
\[
\lambda_{i_0,t}
\ge
\lambda_{i_0}^\star
\exp\left(-1-\frac{\bar V}{\lambda_{i_0}^\star}\right)
>0.
\]
Thus \(\inf_t\|\lambda_t\|_1>0\), and
\[
L(w_t)=\Theta(e^{-\kappa S_t}).
\]

\section{D. Perturbation Argument and Adam/RMSProp Diagnostics}

\paragraph{Deterministic perturbation lemma.}
Let
\[
x_{t+1}=x_t+\eta_t d_t,\qquad
y_{t+1}=y_t+\eta_t(d_t+e_t),
\]
with \(S_t=\sum_{s<t}\eta_s\). If \(\|x_0-y_0\|/S_t\to0\) and
\begin{equation}
\frac1{S_t}\sum_{s<t}\eta_s\|e_s\|\to0,
\label{eq:app-perturb}
\end{equation}
then
\[
\frac{\|x_t-y_t\|}{S_t}\le
\frac{\|x_0-y_0\|}{S_t}
+\frac1{S_t}\sum_{s<t}\eta_s\|e_s\|
\to0 .
\]
This lemma is complete. Its role is to show why the Adam-transfer residual is scientifically relevant.

\paragraph{Adam/RMSProp convention.}
The full-batch Adam update tested in E3 uses
\begin{align*}
m_t&=\beta_1m_{t-1}+(1-\beta_1)\nabla L(w_t),\\
v_t&=\beta_2v_{t-1}+(1-\beta_2)\nabla L(w_t)^{\odot2},
\end{align*}
with optional bias correction and update direction
\[
-\frac{m_t}{\sqrt{v_t}+\eps_t}.
\]
RMSProp is the \(\beta_1=0\) variant under the same denominator convention.

\paragraph{Prespecified transfer residuals.}
E3 records
\begin{align*}
R_m(t)&=
\frac{\|m_t-\nabla L(w_t)\|_2}{\|\nabla L(w_t)\|_2+\eps_t},\\
R_v(t)&=
\frac{\|\sqrt{v_t}-|\nabla L(w_t)|\|_2}
{\|\nabla L(w_t)\|_2+\eps_t},
\end{align*}
\[
R_{\mathrm{dyn}}(t)=
\left\|
\frac{-m_t}{\sqrt{v_t}+\eps_t}
-
\frac{-\nabla L(w_t)}{|\nabla L(w_t)|+\eps_t}
\right\|_2,
\]
and the weighted average \(S_t^{-1}\sum_{s<t}\eta_sR_{\mathrm{dyn}}(s)\). If this averaged residual vanished, the perturbation lemma would transfer the main dynamic theorem to the tested adaptive trajectory. Under the prespecified E3 runs, the residual did not vanish and the adaptive methods followed different paths. Mean final path error was \(0.0886\) for smoothed-sign descent, \(0.8432\) for RMSProp, and \(0.8443\) for Adam. Mean weighted transfer residuals were \(3.8404\) for RMSProp and \(3.8668\) for Adam. The least-bad tested variants still had path errors \(0.4516\) for RMSProp with \(\beta_2=0.9\) and \(0.4226\) for Adam with \(\beta_1=0.9,\beta_2=0.9\) and bias correction. These diagnostics do not support the broad Adam-transfer target under the tested conditions; they do not prove an impossibility result for every adaptive variant.

\begin{figure*}[t]
\centering
\includegraphics[width=.96\textwidth]{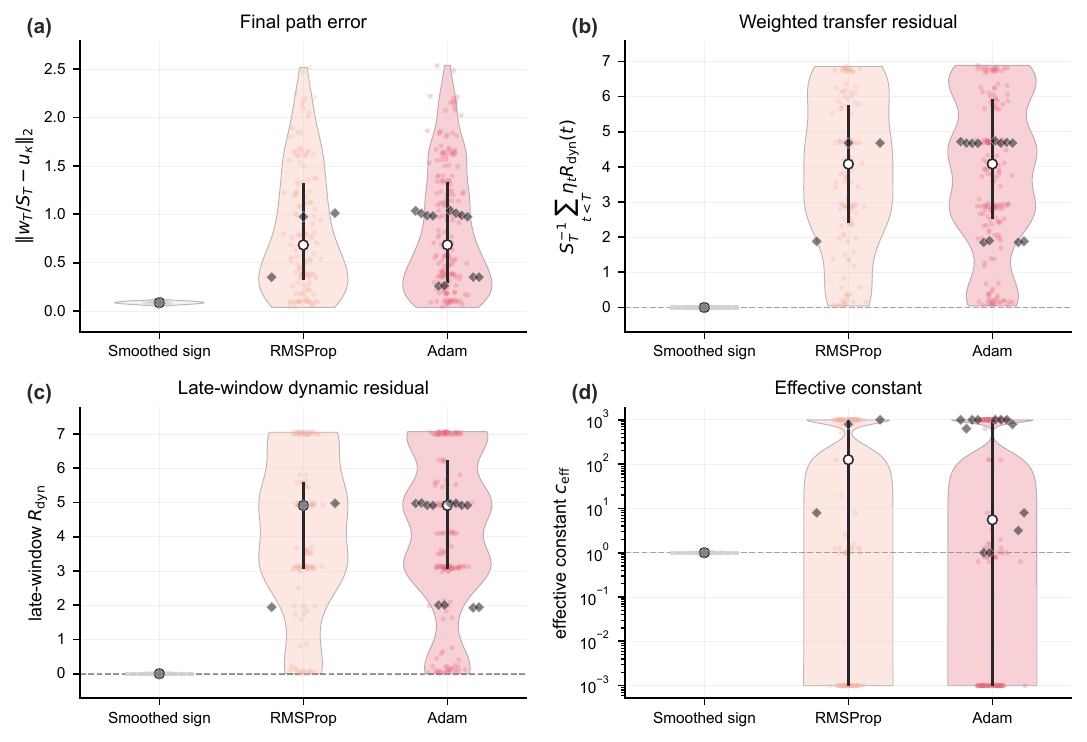}
\caption{Adaptive-method boundary diagnostics. The main paper reports a compact summary; these appendix panels show replicate-level distributions for (a) final path error to \(u_\kappa\), (b) the weighted transfer residual, (c) late-window \(R_{\mathrm{dyn}}\), and (d) the effective-constant diagnostic. Open circles and vertical bars mark medians and interquartile ranges, translucent points are replicate-level configurations, and diamonds mark optimizer-variant medians over the tested \(\beta\) and bias-correction settings. The tested adaptive methods do not satisfy the transfer-residual condition under the prespecified grid; these diagnostics do not establish an impossibility theorem for all adaptive variants.}
\label{fig:app-e3}
\end{figure*}

\section{E. Fixed-\(\epsilon\) Crossover Proof Program}

The transition definition is \eqref{eq:app-tau}. The numerically supported target is
\[
\frac{S_{\tau_\eps(c)}}{\log(1/\eps)}\to\frac1{\ginf},
\]
with \(c\in\{0.5,1,2\}\). The proof program requires two-sided gradient-tail bounds for the fixed-\(\epsilon\) trajectory, uniformly up to the transition window:
\begin{equation}
c_0e^{-\ginf S_t}
\le
\|-\nabla L(w_t)\|_\infty
\le
Ce^{-\ginf S_t}.
\label{eq:app-tail-bounds}
\end{equation}

\paragraph{Conditional upper implication.}
If the upper bound in \eqref{eq:app-tail-bounds} holds, then the threshold is satisfied whenever
\[
Ce^{-\ginf S_t}\le c\eps,
\]
or
\[
S_t\ge \frac1{\ginf}\log\frac1{\eps}
+\frac1{\ginf}\log\frac{C}{c}.
\]
Up to the \(S_t\)-mesh of the schedule, this gives
\[
S_{\tau_\eps(c)}
\le
\frac1{\ginf}\log\frac1{\eps}
+O(1).
\]

\paragraph{Conditional lower implication.}
If the lower bound in \eqref{eq:app-tail-bounds} holds before the transition, then for
\[
S_t\le \frac1{\ginf}\log\frac1{\eps}-M
\]
we have
\[
\|-\nabla L(w_t)\|_\infty
\ge c_0e^{\ginf M}\eps .
\]
Choosing \(M\) so that \(c_0e^{\ginf M}>c\) prevents the transition before that scale and yields
\[
S_{\tau_\eps(c)}
\ge
\frac1{\ginf}\log\frac1{\eps}
-O(1).
\]
The two inequalities imply
\[
S_{\tau_\eps(c)}
=\frac1{\ginf}\log\frac1{\eps}+O(1),
\qquad
\frac{S_{\tau_\eps(c)}}{\log(1/\eps)}\to\frac1{\ginf}.
\]
This is only a conditional implication. The unresolved obligations are a uniform upper tail, a uniform noncanceling lower tail, control of changes in the leading coordinate, pre-transition proximity to the \(\ell_\infty\)-margin geometry, and schedule mesh control. The stronger \(O(\log\log(1/\eps))\) secondary target is not proved.

\begin{figure*}[t]
\centering
\includegraphics[width=.96\textwidth]{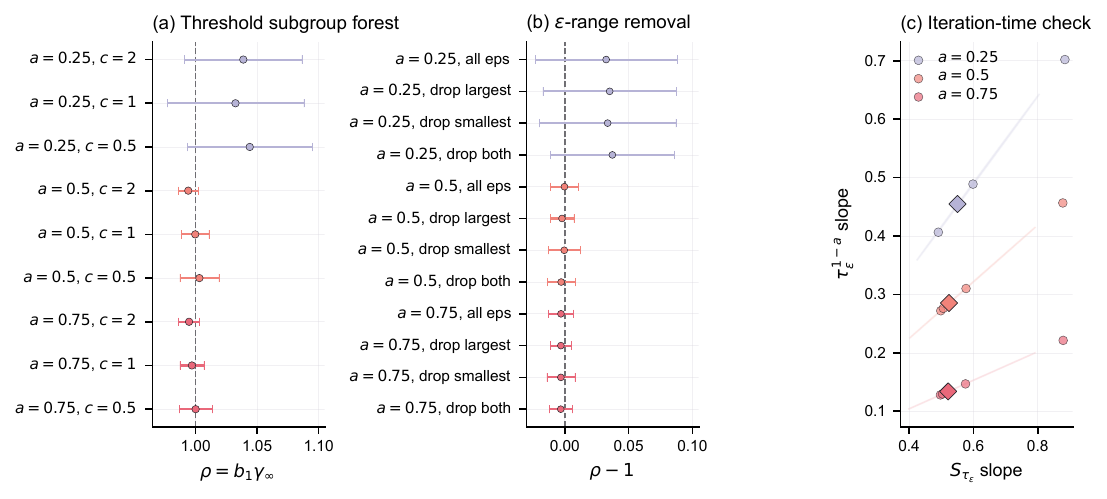}
\caption{Expanded fixed-\(\epsilon\) crossover diagnostics. These panels stress-test the empirical slope relation by threshold subgroup, \(\epsilon\)-range removal, and secondary iteration-time scaling. They are empirical sensitivity checks; the two-sided tail bound remains open.}
\label{fig:app-e4}
\end{figure*}

\section{F. Numerical Solvers and Algorithms}

The core library implements guaranteed-separable synthetic data generation, numerical separability checks, exponential and logistic losses with stable evaluations, cumulative learning-rate schedules, stability schedules, memoryless smoothed-sign dynamics, RMSProp, Adam with optional bias correction, margin solvers, the barrier solver, KKT residuals, and trajectory metrics. All public numerical routines document their mathematical conventions.

\paragraph{Margin solvers.}
The \(\ell_2\)-separator \(v_2\) is solved as a convex quadratic program, and feasibility is checked by the minimum residual of \(Zv-\one\). The \(\ell_\infty\)-margin value is solved over the closed cube \(\|u\|_\infty\le1\), with status and feasibility residuals recorded. Solver failures are written to the processed result files rather than filtered silently.

\paragraph{Barrier solver.}
The theoretical barrier program is open-domain. Numerically, the solver uses \(|u_j|\le1-\delta\) with \(\delta=10^{-7}\), and, where needed for disciplined convex modeling, an auxiliary magnitude variable \(r_j\) satisfying \(r_j\ge u_j\), \(r_j\ge -u_j\), and \(0\le r_j\le1-\delta\). The objective is \(\sum_j[-r_j-\log(1-r_j)]\). KKT residuals include primal feasibility, complementarity, and stationarity residuals computed from \(\nabla B(u)=u/(1-|u|)\). The static program distinguishes solver error from dynamic trajectory error in all summaries.

\paragraph{Data and serialization.}
All experiments use float64 arithmetic and deterministic seeds. The artifact package records configuration hashes, solver statuses, saved trajectory artifacts, and metadata sidecars for all generated figures.

\section{G. Complete Experimental Protocol and Hyperparameter Grids}

The aggregate synthetic datasets use \(n=4d\) signed examples and unfiltered seeds. The two-dimensional dataset used for conceptual path visualization is deliberately labeled illustrative and is excluded from aggregate statistics.

\paragraph{Primary theorem-compatible validation suite.}
The strengthened primary suite uses weighted exponential loss and checks every theorem assumption before including a run in the main Theorem~1 evidence. It uses three synthetic families: isotropic random separable data, controlled-support-vector data with varied support counts, and correlated ill-conditioned data. The aggregate grid is \(d\in\{10,25,50,100\}\), \(n=4d\), twenty unfiltered seeds per setting, \(\kappa/\ginf\in\{0.10,0.25,0.50,0.75,0.90\}\), \(a_i=1\) and bounded log-uniform positive weights, and \(\eta_t=\eta_0(t+t_0)^{-0.75}\) with \(\eta_0=0.5/G\). Secondary schedule checks use powers \(0.60\) and \(0.90\). The main aggregate reports only records with \(0<\kappa<\ginf\), \(S_t\to\infty\), \(\sum_t\eta_t^2<\infty\), \(\eta_tG\le0.5\), successful barrier optimization, and successful trajectory completion.

\paragraph{E1: barrier path recovery.}
E1 uses \(d\in\{10,25,50\}\), \(n=4d\), at least five unfiltered random seeds per configuration, and
\[
\kappa/\ginf\in\{0.10,0.20,0.35,0.50,0.65,0.80,0.90\}.
\]
Reported metrics include path error, margin error, directional error, KKT residual, and gradient exponent diagnostics.

\paragraph{E2: rate diagram and endpoints.}
E2 uses the grid \(0,0.05,\ldots,0.90,0.95,0.99,1.01,1.20,1.50\) times \(\ginf\). It tests the normalized margin diagram, small-\(\kappa\) convergence toward \(v_2\), and near-endpoint convergence toward \(U_\infty\) when the endpoint is unique.

\paragraph{E3: optimizer boundary.}
E3 compares memoryless smoothed-sign dynamics, RMSProp, full Adam, and Adam with and without bias correction. The prespecified grid is \(\beta_1\in\{0,0.9\}\), \(\beta_2\in\{0.9,0.99,0.999\}\), and \(\kappa/\ginf\in\{0.20,0.50,0.65\}\), using the same aggregate dimensions and seeds as E1/E2 where applicable.

\paragraph{E4: fixed-\(\epsilon\) crossover.}
E4 uses \(d=10\), \(\eps\in\{10^{-2},10^{-3},\ldots,10^{-12}\}\), thresholds \(c\in\{0.5,1,2\}\), and learning-rate schedules
\[
\eta_t=\eta_0(t+t_0)^{-a},
\qquad a\in\{0.25,0.50,0.75\}.
\]
The primary fit is \(S_{\tau_\eps}=b_0+b_1\log(1/\eps)\), and the reported slope ratio is \(b_1/(1/\ginf)\). Sensitivity is reported after removing the largest and smallest \(\epsilon\) values.

\paragraph{E5: robustness and failure boundaries.}
E5 tests exponential and logistic losses, multiple schedules aligned by \(S_t\), multiple \(\eps_0\) values, dimensions up to \(d=100\), varying support-vector counts, near-nonunique \(\ell_\infty\) solutions, nearly zero gradient coordinates, data rescaling, coordinate rescaling, and nonzero initialization scales. Outcomes are classified as supported within assumptions, supported outside current assumptions, finite-time ambiguity, numerical failure, or genuine counterexample.

\section{H. Extended Results and Ablations}

Figures promoted to the main paper are not duplicated here unless the appendix version adds information. The following panels are expanded diagnostics relative to the main summaries.

\begin{table}[t]
\centering
\footnotesize
\begin{tabular}{lrrrr}
\toprule
Family & Runs & Within & Outside & Error \\
\midrule
Loss & 12 & 12 & 0 & 0.104 \\
Schedules & 9 & 9 & 0 & 0.097 \\
\(\epsilon_0\) & 18 & 18 & 0 & 0.107 \\
Support vectors & 12 & 12 & 0 & 0.087 \\
High dimension & 2 & 2 & 0 & 0.130 \\
Boundary pathologies & 16 & 0 & 16 & 0.106 \\
\bottomrule
\end{tabular}
\caption{Finite-horizon robustness classification relative to theorem scope. ``Within'' and ``Outside'' are finite-horizon classifications relative to the exact theorem assumptions, not new theorem statuses. No numerical failure or genuine counterexample was observed in the prespecified grid.}
\label{tab:app-robust}
\end{table}

\begin{figure*}[t]
\centering
\includegraphics[width=.96\textwidth]{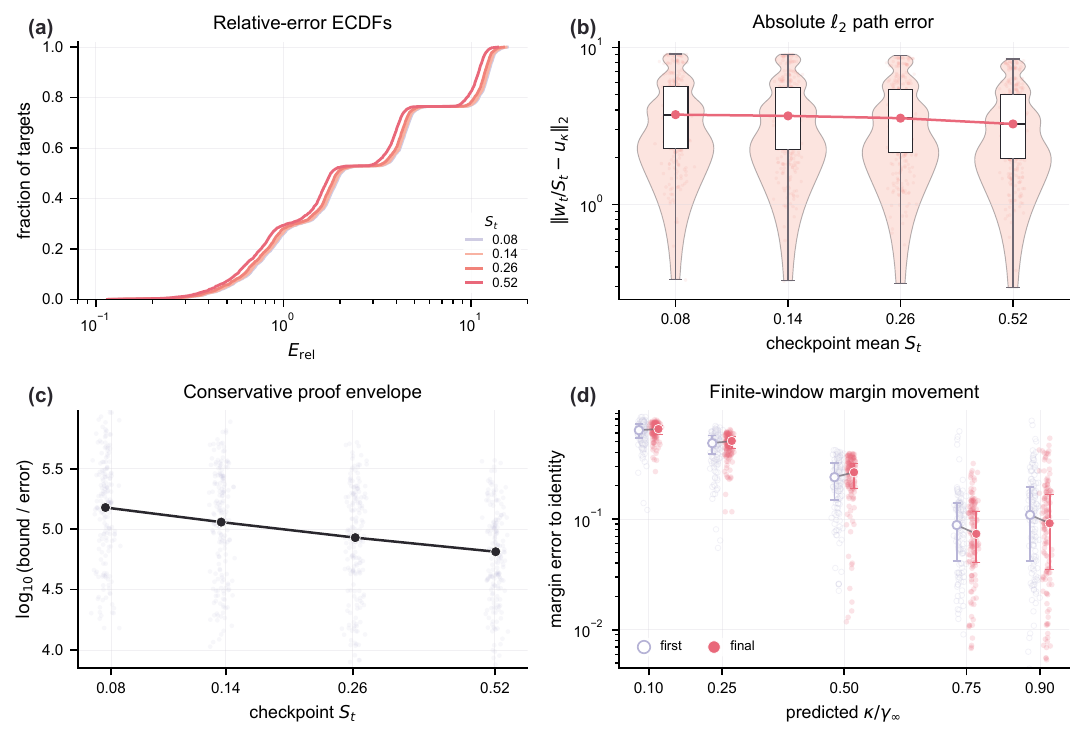}
\caption{Extended Theorem 1 validation diagnostics. These panels expand the main theorem-validation dashboard by showing relative-error ECDFs, absolute-error scales, a conservative certified-envelope quantile trend, and paired first-to-final margin-error movement. They use theorem-compatible weighted exponential-loss records. The certified envelope is not a tight empirical-rate estimate, and the margin diagnostic is not used to claim sharp rate matching.}
\label{fig:app-primary-traces}
\end{figure*}

\begin{figure*}[t]
\centering
\includegraphics[width=.96\textwidth]{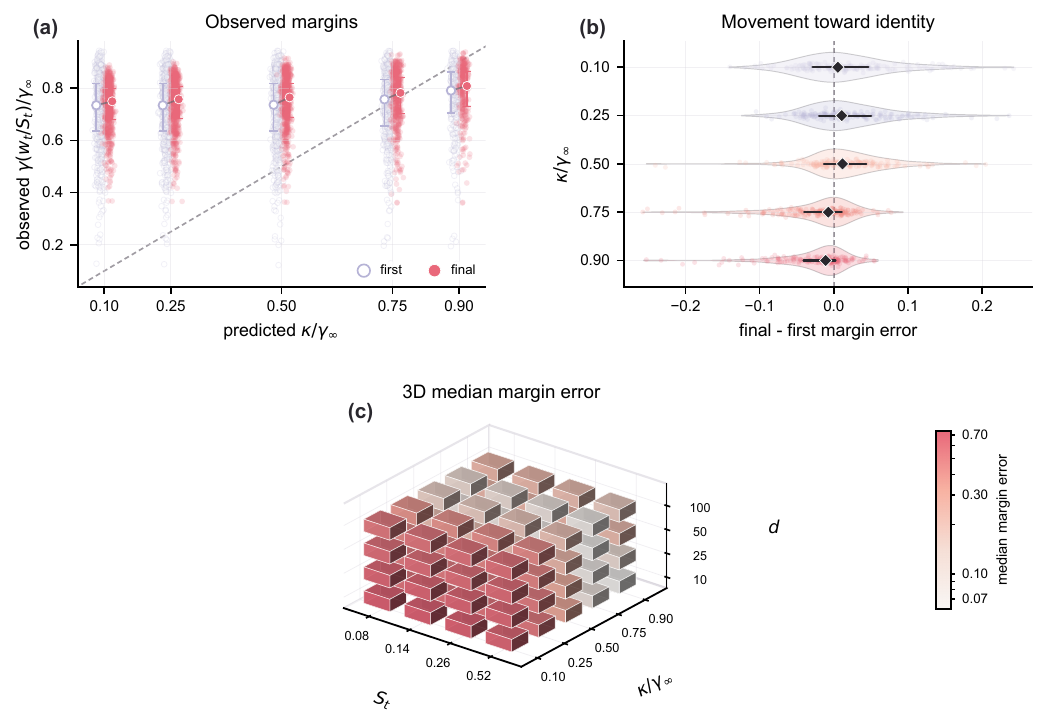}
\caption{Primary-window finite-horizon margin diagnostics. Complementing the main dashboard and the extended validation figure, these panels show replicate-level first/final observed normalized margins, final-minus-first margin-error distributions, and a 3D heatmap of median margin error across checkpoint \(S_t\), rate \(\kappa/\gamma_\infty\), and dimension \(d\). All panels use theorem-compatible weighted exponential-loss records from the primary aggregate suite. The diagnostics describe saved-horizon behavior and are not used to claim sharp margin-rate matching.}
\label{fig:app-envelope-margin}
\end{figure*}

\begin{figure*}[t]
\centering
\includegraphics[width=.96\textwidth]{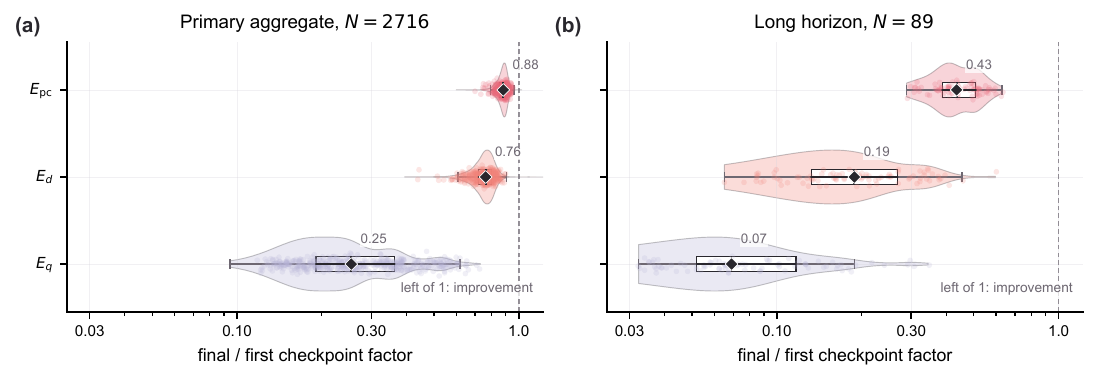}
\caption{Finite-window convergence-factor scorecard. The two panels compare final/first factors for \(E_{\rm pc}\), \(E_d\), and \(E_q\) in (a) the primary aggregate suite and (b) the long-horizon showcase. Values below one indicate movement between saved checkpoints. Median factors and improvement fractions are retained in the figure metadata; the plotted distributions are finite-horizon diagnostics and are not used as sharp-rate claims.}
\label{fig:app-convergence-factors}
\end{figure*}

\FloatBarrier
\end{document}